\pdfoutput=1
\documentclass[11pt]{article}
\usepackage{graphicx}
\usepackage{coling}

\usepackage{times}
\usepackage{latexsym}
\usepackage{booktabs}
\usepackage{array}
\usepackage{xcolor}
\definecolor{numb}{rgb}{0.58,0,0.82} % This is an example RGB value for the color

\usepackage{listings}
\lstdefinelanguage{json}{
  basicstyle=\small\ttfamily,
  numbers=left,
  numberstyle=\tiny,
  stepnumber=1,
  numbersep=8pt,
  showstringspaces=false,
  breaklines=true,
  frame=lines,
  stringstyle=\color{red},
  morestring=[b]",
  literate=
    *{0}{{{\color{numb}0}}}{1}
     {1}{{{\color{numb}1}}}{1}
     {2}{{{\color{numb}2}}}{1}
     {3}{{{\color{numb}3}}}{1}
     {4}{{{\color{numb}4}}}{1}
     {5}{{{\color{numb}5}}}{1}
     {6}{{{\color{numb}6}}}{1}
     {7}{{{\color{numb}7}}}{1}
     {8}{{{\color{numb}8}}}{1}
     {9}{{{\color{numb}9}}}{1}
}

\usepackage[T1]{fontenc}
\usepackage[utf8]{inputenc}

\usepackage{microtype}

\usepackage{inconsolata}

\title{Abstract2Appendix: Academic Reviews Enhance LLM Long-Context Capabilities}

\author{
  Shengzhi Li\textsuperscript{1}, Kittipat Kampa\textsuperscript{1}, Rongyu Lin\textsuperscript{2}, Bohang Li\textsuperscript{3}, \\ 
  \textbf{Shichao Pei}\textsuperscript{4*} \\[1em]
  \textsuperscript{1}TIFIN, Boulder, CO, USA \\
  \textsuperscript{2}Clark University, Worcester, MA, USA \\
  \textsuperscript{3}Shopee, Singapore \\
  \textsuperscript{4}University of Massachusetts Boston, Boston, MA, USA
}
\begin{document}
\maketitle
\begin{abstract}
Large language models (LLMs) have shown remarkable performance across various tasks, yet their ability to handle long-context reading remains challenging. This study explores the effectiveness of leveraging high-quality academic peer review data for fine-tuning LLMs to enhance their long-context capabilities. We compare the Direct Preference Optimization (DPO) method with the Supervised Fine-Tuning (SFT) method, demonstrating DPO's superiority and data efficiency. Our experiments show that the fine-tuned model achieves a 4.04-point improvement over phi-3 and a 2.6\% increase on the Qasper benchmark using only 2000 samples. Despite facing limitations in data scale and processing costs, this study underscores the potential of DPO and high-quality data in advancing LLM performance. Our dataset is on \href{https://github.com/findalexli/Abstract2Appendix}{GitHub}.

Additionally, the zero-shot benchmark results indicate that aggregated high-quality human reviews are overwhelmingly preferred over LLM-generated responses, even for the most capable models like GPT-4o. This suggests that high-quality human reviews are extremely rich in information, reasoning, and long-context retrieval, capabilities that even the most advanced models have not fully captured. These findings highlight the high utility of leveraging human reviews to further advance the field. 
\end{abstract}

\section{Introduction}

Large language models (LLMs) trained on vast amounts of data and computation have achieved remarkable results across tasks \cite{ding2024semantic, tian2024mmrec}. However, as the complexity of tasks these models target increases, their ability to comprehend long texts faces unprecedented challenges. One such challenge is the "Lost in the Middle" problem~\cite{he2023lost}, where models perform well when relevant information is at the beginning or end of the input prompt but significantly decline in performance when the relevant information is in the middle.

Research papers and reviews represent high-quality long-text data, as reviews cover different sections of papers, highlight issues, and provide summaries and abstracts of the content. Numerous efforts to collect and build academic peer review datasets have emerged and been applied to various tasks~\cite{lin2023moprd,gao-etal-2019-rebuttal,10.1371/journal.pone.0259238}. However, current research using these datasets primarily focuses on downstream tasks, such as article acceptance prediction~\cite{kang2018dataset,stappen20_interspeech} and citation relationship prediction~\cite{plank2019citetracked}, with little discussion on their impact on language models.

To address these challenges, we introduce academic peer review data into the fine-tuning process of LLMs.  Based on 2000 parsed pdfs, we conducted experiments using both Supervised Fine-Tuning (SFT) and Direct Preference Optimization (DPO) \cite{rafailov2024direct} methods.

\vspace{1em}

In general, our main contributions are:

\begin{enumerate}
    \item \textbf{Proposing academic reviews as a high-quality long-text supervision dataset}: to our knowledge, we are the first to propose using scientific reviews as a natural source of high-quality supervision signal long-context data, providing a robust dataset for training and evaluating language models.
    \item \textbf{Effective implementation via DPO to add supervision signal}: We conduct a comparative analysis of Supervised Fine-Tuning (SFT) and Direct Preference Optimization (DPO) approaches and found DPO is effective at improving language models' performance on long-text understanding tasks, 
\end{enumerate}

\section{Related Work}

% \begin{figure*}[ht]
%     \centering
%     \includegraphics[width=1\linewidth]{sorted_delta.png}
%     \caption{Sub-task improvement on the LongBech tasks}
%     \label{fig:subtask}
% \end{figure*}
Recent works to improve the long context capabilities of large language models can be divided into the following categories, including but not limited to i) Length Extrapolation. Methods such as RoPE ABF~\cite{xiong2023effective} and Long RoPE~\cite{ding2024longrope} are enhancements of the classic RoPE encoding, while AliBi \cite{press2022train} introduces a novel linear biased position encoding mechanism. ii) Updated or New Model Structures. For example, the decoupled network architecture of LONGMEM \cite{wang2023augmenting} and recurrent transformer \cite{wu2022memformer}. iii) Instruction fine-tuning. Improving the long context capabilities of large language models through fine-tuning based on long text datasets \cite{xiong2023effective,bai2024longalign}. It is worth mentioning that they \cite{xiong2023effective} not only improved the long context capabilities of models by incorporating the loss of predicting input sequences into the loss function but also found a cheaper but more effective pre-training way to make LLM's long context reading ability stronger: starting from shorter context and gradually increasing sequence length for continuous pre-training.

\begin{table*}[t]
\scriptsize % 调整整体字体大小
\centering
\caption{Performance comparison of different LLMs on LongBench. Results marked with * indicate the results reported by LongBench. The number (e.g., '4k') represents the maximum input length that each model can handle.  The table compares performance across various datasets, including NarrativeQA, Qasper, MultiField-en, HotpotQA, 2WikiMQA, Musique, GovReport, QMSum, MultiNews, TREC, TriviaQA, SAMSum, PassageCount, PassageRe, Lcc, RepoBench-P, and their averaged score . Base models Llama-2-7B-chat-4k and phi-3-mini-4B-128k were pre-trained and did not have a fine-tuning data scale reported. While existing reported long-context fine-tuning work improved Llama-2-7B, our efforts excel and improved the average score of phi-3 by 4.04 by using only 2000 samples.}
\label{tab:longbench}
\begin{tabular*}{\textwidth}{@{\extracolsep{\fill}}p{2.2cm}p{0.6cm}p{0.35cm}p{0.35cm}p{0.35cm}p{0.35cm}p{0.35cm}p{0.35cm}p{0.35cm}p{0.35cm}p{0.35cm}p{0.35cm}p{0.35cm}p{0.35cm}p{0.35cm}p{0.35cm}p{0.35cm}p{0.35cm}p{0.6cm}p{2.2cm}}
\toprule
LLMs & \rotatebox{60}{Fine-tuning Data} & \rotatebox{60}{NarrativeQA} & \rotatebox{60}{Qasper} & \rotatebox{60}{MultiField-en} & \rotatebox{60}{HotpotQA} & \rotatebox{60}{2WikiMQA} & \rotatebox{60}{Musique} & \rotatebox{60}{GovReport} & \rotatebox{60}{QMSum} & \rotatebox{60}{MultiNews} & \rotatebox{60}{TREC} & \rotatebox{60}{TriviaQA} & \rotatebox{60}{SAMSum} & \rotatebox{60}{PassageCount} & \rotatebox{60}{PassageRe} & \rotatebox{60}{Lcc} & \rotatebox{60}{RepoBench-P} & \rotatebox{60}{Avg. Score} \\
\midrule
Llama-2-7B-chat-4k* &  & 18.7 & 19.2 & 36.8 & 34.34 & 34.2 & 14.13 & 27.3 & 21.35 & 25.8 & 61.5 & 77.8 & 40.7 & 2.1 & 9.8 & 52.4 & 43.8 & 32.50 \\
LongChat1.5-7B-32k* & 18k & 16.9 & 27.7 & 41.4 & 31.5 & 40.6 & 9.7 & 30 & 21.9 & 24.3 & 55.3 & 82.3 & 34 & 1 & 3 & 53 & 55.3 & 32.99 \\
together/llama-2-7b-32k* & 38k & 15.5 & 26.3 & 43.1 & 32.36 & 41.3 & 6.9 & 25.3 & 22.5 & 22.9 & 57.2 & 87.7 & 43.7 & 1 & 2.3 & 63.7 & 61.7 & 34.60 \\
CLEX-7B-16k* & 3.9M & 19.1 & 28.2 & 44.6 & 28.44 & 42.6 & 14.6 & 31.6 & 32.5 & 22.7 & 55.5 & 84.9 & 42.8 & 0 & 11.5 & 59 & 56.8 & 35.93 \\
CodeLLaMA-7B-16k* & 0.5T Token & 17.6 & 24.4 & 43.9 & 34.47 & 41.4 & 14.2 & 32 & 27.5 & 23.9 & 61.3 & 84.9 & 47.2 & 1.1 & 1.3 & 64.3 & 56.5 & 36.00 \\
Vicuna1.5-7B-16k* & 125k & 16.9 & 27.5 & 41.4 & 28.1 & 39.4 & 8.7 & 30 & 21.2 & 20.1 & 53.3 & 86.3 & 44 & 3.1 & 7 & 60.1 & 44 & 33.19 \\
phi-3-mini-4B-128k  &  & 21.82 & 37.44 & 49.79 & 46.34 & 36.89 & 27.49 & 32.72 & 22.69 & 24.49 & 8.5 & 82.03 & 32.41 & 2.7 & 68 & 3.55 & 5.1 & 31.37 \\
phi-3-mini-4B-128k-dpo(Ours) & 2k & 21.71 & 38.97 & 50.25 & 45.84 & 38.92 & 26.97 & 31.84 & 22.71 & 24.95 & 26.5 & 84.13 & 35.36 & 2.25 & 76.5 & 20.2 & 19.51 & 35.41 \\
\bottomrule
\end{tabular*}
\end{table*}

\begin{table*}[ht]
\centering
\caption{Model Win Rates Judged by Gemini pro and Mixtral-8x22B}
\begin{tabular}{lrr}
\toprule
Model & Win rate (Gemini pro as judge) & Win rate (Mixtral-8x22B as judge)\\
\midrule
GPT-3.5-turbo & 0.0 & 30.9 \\
GPT4-turbo-0409 & 33.0 & 28.9 \\
GPT-4o & 51.5 & 36.1 \\
Mistral-7B-Instruct-v0.3 & 11.3 & 7.2 \\
Mixtral-8x7B-Instruct-v0.1 & 27.1 & 14.4 \\
Qwen1.5-1.8B-Chat & 0.0 & 22.7 \\
Qwen1.5-4B-Chat & 0.0 & 15.5 \\
Qwen1.5-7B-Chat & 18.6 & 14.4 \\
Qwen2-72B-Instruct & 35.1 & 9.3 \\
\bottomrule
\end{tabular}
\label{tab:model_win_rates}
\end{table*}

\section{Methods}

\subsection{Data Set Preparation}
We download the ICLR 2024 submitted papers, including the PDFs and the corresponding reviews. We use Amazon Textract to extract tables where needed and convert the PDF file into an HTML text file. 

Given that each paper has 3-6 reviews from ICLR reviewers, we use GPT-4 to generate an aggregated review across the helpful, attention-to-detailed portion of each review, which we refer to as the aggregated review. We prompt the LLM not to rely on its knowledge but only to help aggregate the answer coherently.  

% The process of converting a PDF document to an HTML file involves several systematic steps, leveraging various tools and AWS services for efficiency and scalability. Initially, the PDF is analyzed to extract metadata, such as the total number of pages. The PDF is then split into individual pages, and each page is converted into a JPEG image. These pages are processed in parallel batches to manage memory and optimize performance effectively. Once the images are generated, they are processed using Amazon Textract to extract structured data from each page, including texts, layouts, tables, forms, and signatures. This data is serialized and stored locally. Next, concurrent processing is employed to expedite the extraction process, utilizing multiple threads to handle multiple pages simultaneously. The extracted data is then converted to HTML format, with the output of each page written to an individual HTML file. These HTML files from the same document are subsequently combined into a single HTML document. Additional formatting is applied to enhance readability, such as replacing newlines with line breaks and adding borders to tables. This automated pipeline ensures a robust and efficient conversion of PDF documents to a web-friendly HTML format, facilitating easy access and review.

\begin{figure}
    \centering
    \includegraphics[width=1\linewidth]{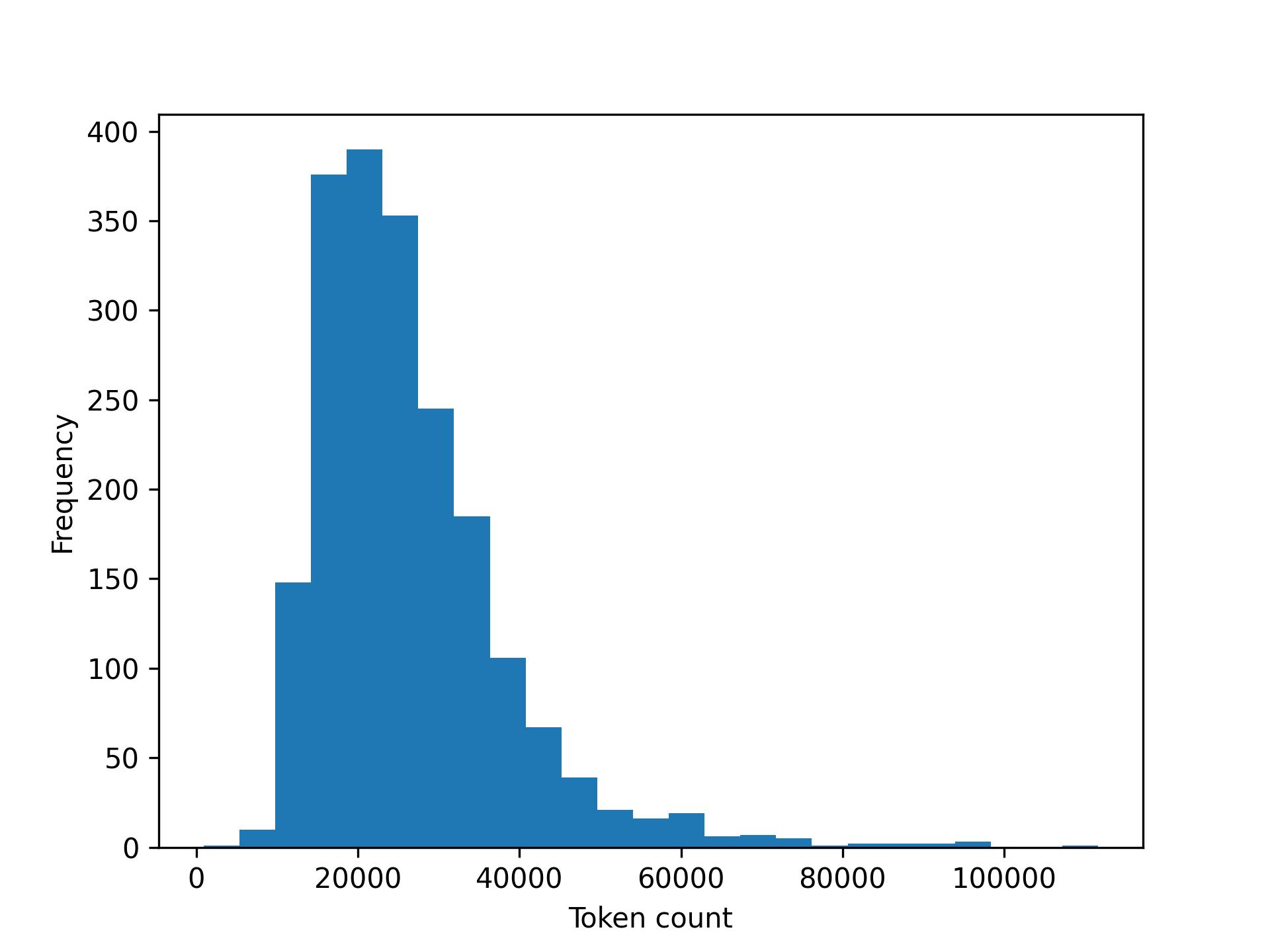}
    \caption{Token count distribution for combinations of paper and review (n=2005), with an average of 26353 tokens and standard deviations of 11774. In downstream experiments, we filter out samples longer than 33000}
    \label{fig:token_distribution}
\end{figure}

\subsection{Zero-shot benchmarks}
We would like to assess existing models' capabilities in generating the response. We employed the following models with long context limit: GPT-3.5 and GPT4 series \cite{openai2024gpt4}, Mistral and Mixtral \cite{jiang2023mistral}, and Qwen \cite{qwen}. We note GPT-3.5-16k would not see entire paper. The following sentence instructions are added: "Given the following paper, help write a review for the paper. The review should be helpful and constructive and should point out specific parts of the paper that need attention" before the paper string. We note that 

We use the MT-Bench system prompt as a judge idea \cite{zheng2023judging}, and chose Mixtral-8x22b-Instuct (v0.1) \cite{mistral2024}and Gemini-pro-1.5 \cite{geminiteam2024gemini} due to its high performance and the fact that it could serve as a less biased judge than GPT-4o which generated our aggregated data.  To counter the positional bias, we randomly flip the position of the candidate's answer with the aggregated review answer.

\subsection{Fine-tuning}
We would like to assess whether high-quality scientific reviews can help add additional supervision signals to the models and thus experiment to fine-tune the model.

We chose Phi-3-mini-128k \cite{abdin2024phi3} for its small size and 128K context length. This model, an extended context version of phi-3-mini, retains the high performance of the original 4K context version while adeptly handling longer context tasks. The extension to 128K context length was achieved through a two-stage process: an initial long-context mid-training, followed by mixed long-short post-training with supervised fine-tuning (SFT) and direct preference optimization (DPO).

For the DPO experiment, we employed GPT-4 to rate each review and generate the aggregated review. We select the aggregated review as the preferred response and the lowest-rated review as the rejected review. By optimizing the model's strategy and constructing our loss function, we directly impacted the models' capabilities in generating the response. For the SFT experiment, the task would be to predict the aggregated review. We took the papers’ content as input for our model and asked the model to predict the aggregated review. We fine-tune our model by calculating loss on the review tokens and excluding the context tokens from loss calculation. 

The training was conducted using Deepspeed Stage-3 on a 4x A100 80GB GPU machine with LoRA for parameter-efficient fine-tuning. Experiment parameters are detailed in the appendix \ref{tab:Modelparameters}
\subsection{Long-context Benchmark}
We implemented two standard long-context reasoning and knowledge retrieval benchmarks, including Qasper \cite{dasigi-etal-2021-dataset}, a question-answering reasoning benchmark of NLP papers implemented by lm-evaluation-harness\cite{eval-harness}, and LongBench \cite{bai2023longbench}, a comprehensive long-context LLM benchmark implemented by \cite{2023opencompass}. See the appendix for their description. We included prior model's results as reported, including Llama-2-7B-chat-4k \cite{touvron2023llama}, LongChat \cite{longchat2023}, Together-llama-2-7b-32k \cite{togetherai2024}, Clex-7b-16k \cite{damonlpsg2023clex} and Phi-3-mini-4B-128k (instruct version) \cite{abdin2024phi3}.

\section{Results}

The performance comparison of different LLMs on the LongBench dataset, as presented in Table \ref{tab:longbench}, reveals several key observations. Our fine-tuned phi-3-mini-128k-dpo model consistently demonstrates superior performance across multiple tasks despite utilizing only 2K data for fine-tuning, improving the averaged score on phi-3-mini-4B-128k by 4.04. Specifically, the phi-3-mini-128k-dpo excels in tasks such as Qasper, MultiField-en, HotpotQA, and Lcc. These results highlight the effectiveness of using high-quality academic peer review data for fine-tuning, demonstrating that even limited data can significantly improve long-context reading abilities.

This efficiency contrasts sharply with other models focused on fine-tuning LLama/LLama-2-7B, such as LongChat1.5-7B-32k and together/llama-2-7b-32k, which require significantly larger fine-tuning datasets (18k and 38k, respectively) yet achieve comparable or lower performance delta from the base-model. Vicuna and CLEX would take even more data and compute (125K and 3.9M rows of samples). As long-context fine-tuning is computationally intensive and costly, requiring A100s or H100s GPUs with 80/96GB memory, our method is comparably performant while operating at orders of magnitude smaller compute scale. 

In terms of judging LLM's attempts to review the paper, we computed the win rate which indicates how often the LLM's response was preferred by the judge over the aggregated human review. Table \ref{tab:model_win_rates} shows that while the most capable LLMs are increasingly competent, the aggregated human reviews are largely preferred. This highlights the richness of high-quality human reviews regarding information, reasoning, and long-context retrieval—capabilities that current LLMs struggle to fully replicate. The trend of larger models, like those within the Qwen family or Mistral family of models, shows improved but still insufficient performance. 

\section{Discussion}
\begin{table}[t]
\centering
\caption{Effect of data scaling on the DPO model from 600 to 2000 samples compared to the baseline model on the Qasper benchmark.}
\begin{tabular}{|l|c|c|}
\hline
\textbf{Model} & \textbf{Dataset Size} & \textbf{F1 Score} \\ \hline
Base Model     & N/A                   & 0.640 \\ \hline
DPO      & 600                   & 0.645 \\ \hline
DPO      & 2000                  & 0.666 \\ \hline
\end{tabular}
\label{table:data_scaling}
\end{table}

\begin{table}[t]
\centering
\caption{Comparison of SFT and DPO methods at 600 samples demonstrating the superiority of the DPO method on the Qasper benchmark.}
\begin{tabular}{|l|c|c|}
\hline
\textbf{Model} & \textbf{F1 Score} \\ \hline
SFT & 0.585 \\ \hline
DPO & 0.666 \\ \hline
\end{tabular}
\label{table:method_comparison}
\end{table}
\subsection{Effect of data scaling}
We used 2000 samples to fine-tune the DPO model. Compared to the baseline model and the DPO model fine-tuned with 600 samples, the results are highly encouraging. As shown in Table \ref{table:data_scaling}, the F1 score on the Qasper dataset improved by 0.5\% when the baseline model was fine-tuned with 600 samples. This improvement rose to 2.5\% when the sample size increased to 2000. These findings highlight the positive impact of data scaling on the DPO model’s performance. 
\subsection{Effect of SFT versus DPO}

In our initial experiments using the SFT method, we observed that fine-tuning with the SFT method on a dataset with 600 samples resulted in a 6\% reduction in the F1 score on the Qasper dataset. In contrast, using the same number of samples, the DPO method improved the F1 score by 0.5\%. This indicates that for few-shot learning, the DPO method is more effective in fine-tuning LLMs than the SFT method. This is consistent with our previous finding on multi-modal LLM benefits more from DPO than SFT \cite{li2024multi}. 
\section{Conclusion}
This study highlights the effectiveness of leveraging high-quality academic peer review data for improving the long-context capabilities of large language models (LLMs). Our results demonstrate the superiority of the Direct Preference Optimization (DPO) method over the Supervised Fine-Tuning(SFT). The most advanced models have not fully captured fine-tuned phi-3-mini-128k-dpo model outperformed phi-3 by 4.04 points, and Qasper by 2.6\% with only 2000 samples. 

The zero-shot benchmark results highlight the limitations of LLMs, including GPT-4o, in replicating the depth and quality of human reviews, particularly in information richness, reasoning, and long-context retrieval. These findings underscore the importance of using high-quality human reviews to improve language models.

\section*{Limitations}
Our bottleneck was the data scale in this study. While Textract offers low latency speed, the cost structure of Textract is high - it costs us \$5200 in AWS credits to process 3000 PDF documents, preventing us from further increasing our scale and diversity of paper sources. We need further work to investigate how increasing our data size can impact our performance. 

Much of a science paper's signal comes from visual signals, including graphs, charts, and plots.  Future work will incorporate tools that can handle both text and graphical content for a more comprehensive analysis of scientific documents \cite{li2023scigraphqa}. From the data parsing side, we explored Unstructured.io's potential to create a multi-modal document including figures and included a comparison of Textract, Unstructured.io, and GPT-4V, including output format and processing cost, is in Appendix \ref{sec:appendix-compare-tools}, while admittedly the service's low reliability prevented us from pursuing. From the modeling side, future work needs to employ multi-modal LLM and consider the long context from a multi-modal perspective. 

\bibliography{anthology,custom}

\begin{thebibliography}{33}
\providecommand{\natexlab}[1]{#1}

\bibitem[{Abdin et~al.(2024)Abdin, Jacobs, Awan, Aneja, Awadallah, Awadalla, Bach, Bahree, Bakhtiari, Bao, Behl, Benhaim, Bilenko, Bjorck, Bubeck, Cai, Cai, Mendes, Chen, Chaudhary, Chen, Chen, Chen, Chen, Chopra, Dai, Giorno, de~Rosa, Dixon, Eldan, Fragoso, Iter, Gao, Gao, Gao, Garg, Goswami, Gunasekar, Haider, Hao, Hewett, Huynh, Javaheripi, Jin, Kauffmann, Karampatziakis, Kim, Khademi, Kurilenko, Lee, Lee, Li, Li, Liang, Liden, Liu, Liu, Liu, Lin, Lin, Luo, Madan, Mazzola, Mitra, Modi, Nguyen, Norick, Patra, Perez-Becker, Portet, Pryzant, Qin, Radmilac, Rosset, Roy, Ruwase, Saarikivi, Saied, Salim, Santacroce, Shah, Shang, Sharma, Shukla, Song, Tanaka, Tupini, Wang, Wang, Wang, Wang, Ward, Wang, Witte, Wu, Wyatt, Xiao, Xu, Xu, Xu, Yadav, Yang, Yang, Yang, Yang, Yu, Yuan, Zhang, Zhang, Zhang, Zhang, Zhang, Zhang, Zhang, and Zhou}]{abdin2024phi3}
Marah Abdin, Sam~Ade Jacobs, Ammar~Ahmad Awan, Jyoti Aneja, Ahmed Awadallah, Hany Awadalla, Nguyen Bach, Amit Bahree, Arash Bakhtiari, Jianmin Bao, Harkirat Behl, Alon Benhaim, Misha Bilenko, Johan Bjorck, Sébastien Bubeck, Qin Cai, Martin Cai, Caio César~Teodoro Mendes, Weizhu Chen, Vishrav Chaudhary, Dong Chen, Dongdong Chen, Yen-Chun Chen, Yi-Ling Chen, Parul Chopra, Xiyang Dai, Allie~Del Giorno, Gustavo de~Rosa, Matthew Dixon, Ronen Eldan, Victor Fragoso, Dan Iter, Mei Gao, Min Gao, Jianfeng Gao, Amit Garg, Abhishek Goswami, Suriya Gunasekar, Emman Haider, Junheng Hao, Russell~J. Hewett, Jamie Huynh, Mojan Javaheripi, Xin Jin, Piero Kauffmann, Nikos Karampatziakis, Dongwoo Kim, Mahoud Khademi, Lev Kurilenko, James~R. Lee, Yin~Tat Lee, Yuanzhi Li, Yunsheng Li, Chen Liang, Lars Liden, Ce~Liu, Mengchen Liu, Weishung Liu, Eric Lin, Zeqi Lin, Chong Luo, Piyush Madan, Matt Mazzola, Arindam Mitra, Hardik Modi, Anh Nguyen, Brandon Norick, Barun Patra, Daniel Perez-Becker, Thomas Portet, Reid Pryzant, Heyang
  Qin, Marko Radmilac, Corby Rosset, Sambudha Roy, Olatunji Ruwase, Olli Saarikivi, Amin Saied, Adil Salim, Michael Santacroce, Shital Shah, Ning Shang, Hiteshi Sharma, Swadheen Shukla, Xia Song, Masahiro Tanaka, Andrea Tupini, Xin Wang, Lijuan Wang, Chunyu Wang, Yu~Wang, Rachel Ward, Guanhua Wang, Philipp Witte, Haiping Wu, Michael Wyatt, Bin Xiao, Can Xu, Jiahang Xu, Weijian Xu, Sonali Yadav, Fan Yang, Jianwei Yang, Ziyi Yang, Yifan Yang, Donghan Yu, Lu~Yuan, Chengruidong Zhang, Cyril Zhang, Jianwen Zhang, Li~Lyna Zhang, Yi~Zhang, Yue Zhang, Yunan Zhang, and Xiren Zhou. 2024.
\newblock \href {https://arxiv.org/abs/2404.14219} {Phi-3 technical report: A highly capable language model locally on your phone}.
\newblock \emph{Preprint}, arXiv:2404.14219.

\bibitem[{AI(2024{\natexlab{a}})}]{mistral2024}
Mistral AI. 2024{\natexlab{a}}.
\newblock Mixtral-8x22b-instruct-v0.1.
\newblock \url{https://huggingface.co/mistralai/Mixtral-8x22B-Instruct-v0.1}.
\newblock Accessed: 2024-06-15.

\bibitem[{AI(2024{\natexlab{b}})}]{togetherai2024}
Together AI. 2024{\natexlab{b}}.
\newblock Introducing llama-2-7b-32k: Breaking the limits of context length.
\newblock \url{https://www.together.ai/blog/llama-2-7b-32k}.
\newblock Accessed: 2024-06-15.

\bibitem[{Bai et~al.(2023{\natexlab{a}})Bai, Bai, Chu, Cui, Dang, Deng, Fan, Ge, Han, Huang, Hui, Ji, Li, Lin, Lin, Liu, Liu, Lu, Lu, Ma, Men, Ren, Ren, Tan, Tan, Tu, Wang, Wang, Wang, Wu, Xu, Xu, Yang, Yang, Yang, Yang, Yao, Yu, Yuan, Yuan, Zhang, Zhang, Zhang, Zhang, Zhou, Zhou, Zhou, and Zhu}]{qwen}
Jinze Bai, Shuai Bai, Yunfei Chu, Zeyu Cui, Kai Dang, Xiaodong Deng, Yang Fan, Wenbin Ge, Yu~Han, Fei Huang, Binyuan Hui, Luo Ji, Mei Li, Junyang Lin, Runji Lin, Dayiheng Liu, Gao Liu, Chengqiang Lu, Keming Lu, Jianxin Ma, Rui Men, Xingzhang Ren, Xuancheng Ren, Chuanqi Tan, Sinan Tan, Jianhong Tu, Peng Wang, Shijie Wang, Wei Wang, Shengguang Wu, Benfeng Xu, Jin Xu, An~Yang, Hao Yang, Jian Yang, Shusheng Yang, Yang Yao, Bowen Yu, Hongyi Yuan, Zheng Yuan, Jianwei Zhang, Xingxuan Zhang, Yichang Zhang, Zhenru Zhang, Chang Zhou, Jingren Zhou, Xiaohuan Zhou, and Tianhang Zhu. 2023{\natexlab{a}}.
\newblock Qwen technical report.
\newblock \emph{arXiv preprint arXiv:2309.16609}.

\bibitem[{Bai et~al.(2024)Bai, Lv, Zhang, He, Qi, Hou, Tang, Dong, and Li}]{bai2024longalign}
Yushi Bai, Xin Lv, Jiajie Zhang, Yuze He, Ji~Qi, Lei Hou, Jie Tang, Yuxiao Dong, and Juanzi Li. 2024.
\newblock Longalign: A recipe for long context alignment of large language models.
\newblock \emph{arXiv preprint arXiv:2401.18058}.

\bibitem[{Bai et~al.(2023{\natexlab{b}})Bai, Lv, Zhang, Lyu, Tang, Huang, Du, Liu, Zeng, Hou, Dong, Tang, and Li}]{bai2023longbench}
Yushi Bai, Xin Lv, Jiajie Zhang, Hongchang Lyu, Jiankai Tang, Zhidian Huang, Zhengxiao Du, Xiao Liu, Aohan Zeng, Lei Hou, Yuxiao Dong, Jie Tang, and Juanzi Li. 2023{\natexlab{b}}.
\newblock \href {https://arxiv.org/abs/2308.14508} {Longbench: A bilingual, multitask benchmark for long context understanding}.
\newblock \emph{Preprint}, arXiv:2308.14508.

\bibitem[{Chen et~al.(2023)Chen, Li, Meng, Liang, and Bing}]{damonlpsg2023clex}
Guanzheng Chen, Xin Li, Zaiqiao Meng, Shangsong Liang, and Lidong Bing. 2023.
\newblock \href {https://arxiv.org/abs/2310.16450} {Clex: Continuous length extrapolation for large language models}.
\newblock \emph{arXiv preprint arXiv:2310.16450}.

\bibitem[{Contributors(2023)}]{2023opencompass}
OpenCompass Contributors. 2023.
\newblock Opencompass: A universal evaluation platform for foundation models.
\newblock \url{https://github.com/open-compass/opencompass}.

\bibitem[{Dasigi et~al.(2021)Dasigi, Lo, Beltagy, Cohan, Smith, and Gardner}]{dasigi-etal-2021-dataset}
Pradeep Dasigi, Kyle Lo, Iz~Beltagy, Arman Cohan, Noah~A. Smith, and Matt Gardner. 2021.
\newblock \href {https://doi.org/10.18653/v1/2021.naacl-main.365} {A dataset of information-seeking questions and answers anchored in research papers}.
\newblock In \emph{Proceedings of the 2021 Conference of the North American Chapter of the Association for Computational Linguistics: Human Language Technologies}, pages 4599--4610, Online. Association for Computational Linguistics.

\bibitem[{Ding et~al.(2024{\natexlab{a}})Ding, Zhang, Zhang, Xu, Shang, Xu, Yang, and Yang}]{ding2024longrope}
Yiran Ding, Li~Lyna Zhang, Chengruidong Zhang, Yuanyuan Xu, Ning Shang, Jiahang Xu, Fan Yang, and Mao Yang. 2024{\natexlab{a}}.
\newblock \href {https://arxiv.org/abs/2402.13753} {Longrope: Extending llm context window beyond 2 million tokens}.
\newblock \emph{Preprint}, arXiv:2402.13753.

\bibitem[{Ding et~al.(2024{\natexlab{b}})Ding, Tian, Wang, Zhao, and Li}]{ding2024semantic}
Zhicheng Ding, Jiahao Tian, Zhenkai Wang, Jinman Zhao, and Siyang Li. 2024{\natexlab{b}}.
\newblock Semantic understanding and data imputation using large language model to accelerate recommendation system.
\newblock \emph{arXiv preprint arXiv:2407.10078}.

\bibitem[{Gao et~al.(2023)Gao, Tow, Abbasi, Biderman, Black, DiPofi, Foster, Golding, Hsu, Le~Noac'h, Li, McDonell, Muennighoff, Ociepa, Phang, Reynolds, Schoelkopf, Skowron, Sutawika, Tang, Thite, Wang, Wang, and Zou}]{eval-harness}
Leo Gao, Jonathan Tow, Baber Abbasi, Stella Biderman, Sid Black, Anthony DiPofi, Charles Foster, Laurence Golding, Jeffrey Hsu, Alain Le~Noac'h, Haonan Li, Kyle McDonell, Niklas Muennighoff, Chris Ociepa, Jason Phang, Laria Reynolds, Hailey Schoelkopf, Aviya Skowron, Lintang Sutawika, Eric Tang, Anish Thite, Ben Wang, Kevin Wang, and Andy Zou. 2023.
\newblock \href {https://doi.org/10.5281/zenodo.10256836} {A framework for few-shot language model evaluation}.

\bibitem[{Gao et~al.(2019)Gao, Eger, Kuznetsov, Gurevych, and Miyao}]{gao-etal-2019-rebuttal}
Yang Gao, Steffen Eger, Ilia Kuznetsov, Iryna Gurevych, and Yusuke Miyao. 2019.
\newblock \href {https://doi.org/10.18653/v1/N19-1129} {Does my rebuttal matter? insights from a major {NLP} conference}.
\newblock In \emph{Proceedings of the 2019 Conference of the North {A}merican Chapter of the Association for Computational Linguistics: Human Language Technologies, Volume 1 (Long and Short Papers)}, pages 1274--1290, Minneapolis, Minnesota. Association for Computational Linguistics.

\bibitem[{Ghosal et~al.(2022)Ghosal, Kumar, Bharti, and Ekbal}]{10.1371/journal.pone.0259238}
Tirthankar Ghosal, Sandeep Kumar, Prabhat~Kumar Bharti, and Asif Ekbal. 2022.
\newblock \href {https://doi.org/10.1371/journal.pone.0259238} {Peer review analyze: A novel benchmark resource for computational analysis of peer reviews}.
\newblock \emph{PLOS ONE}, 17(1):1--29.

\bibitem[{He et~al.(2023)He, Pan, Dong, Song, Liu, Liang, Wang, Sun, Zhang, Xie, and Zhang}]{he2023lost}
Junqing He, Kunhao Pan, Xiaoqun Dong, Zhuoyang Song, Yibo Liu, Yuxin Liang, Hao Wang, Qianguo Sun, Songxin Zhang, Zejian Xie, and Jiaxing Zhang. 2023.
\newblock \href {https://arxiv.org/abs/2311.09198} {Never lost in the middle: Improving large language models via attention strengthening question answering}.
\newblock \emph{Preprint}, arXiv:2311.09198.

\bibitem[{Jiang et~al.(2023)Jiang, Sablayrolles, Mensch, Bamford, Chaplot, de~las Casas, Bressand, Lengyel, Lample, Saulnier, Lavaud, Lachaux, Stock, Scao, Lavril, Wang, Lacroix, and Sayed}]{jiang2023mistral}
Albert~Q. Jiang, Alexandre Sablayrolles, Arthur Mensch, Chris Bamford, Devendra~Singh Chaplot, Diego de~las Casas, Florian Bressand, Gianna Lengyel, Guillaume Lample, Lucile Saulnier, Lélio~Renard Lavaud, Marie-Anne Lachaux, Pierre Stock, Teven~Le Scao, Thibaut Lavril, Thomas Wang, Timothée Lacroix, and William~El Sayed. 2023.
\newblock \href {https://arxiv.org/abs/2310.06825} {Mistral 7b}.
\newblock \emph{Preprint}, arXiv:2310.06825.

\bibitem[{Kang et~al.(2018)Kang, Ammar, Dalvi, van Zuylen, Kohlmeier, Hovy, and Schwartz}]{kang2018dataset}
Dongyeop Kang, Waleed Ammar, Bhavana Dalvi, Madeleine van Zuylen, Sebastian Kohlmeier, Eduard Hovy, and Roy Schwartz. 2018.
\newblock \href {https://doi.org/10.18653/v1/N18-1149} {A dataset of peer reviews ({P}eer{R}ead): Collection, insights and {NLP} applications}.
\newblock In \emph{Proceedings of the 2018 Conference of the North {A}merican Chapter of the Association for Computational Linguistics: Human Language Technologies, Volume 1 (Long Papers)}, pages 1647--1661, New Orleans, Louisiana. Association for Computational Linguistics.

\bibitem[{Li et~al.(2023)Li, Shao, Xie, Sheng, Zheng, Gonzalez, Stoica, Ma, and Zhang}]{longchat2023}
Dacheng Li, Rulin Shao, Anze Xie, Ying Sheng, Lianmin Zheng, Joseph Gonzalez, Ion Stoica, Xuezhe Ma, and Hao Zhang. 2023.
\newblock How long can context length of open-source llms truly promise?

\bibitem[{Li et~al.(2024)Li, Lin, and Pei}]{li2024multi}
Shengzhi Li, Rongyu Lin, and Shichao Pei. 2024.
\newblock \href {https://doi.org/10.18653/v1/2024.acl-long.765} {Multi-modal preference alignment remedies degradation of visual instruction tuning on language models}.
\newblock In \emph{Proceedings of the 62nd Annual Meeting of the Association for Computational Linguistics (Volume 1: Long Papers)}, pages 14188--14200, Bangkok, Thailand. Association for Computational Linguistics.

\bibitem[{Li and Tajbakhsh(2023)}]{li2023scigraphqa}
Shengzhi Li and Nima Tajbakhsh. 2023.
\newblock Scigraphqa: A large-scale synthetic multi-turn question-answering dataset for scientific graphs.
\newblock \emph{arXiv preprint arXiv:2308.03349}.

\bibitem[{Lin et~al.(2023)Lin, Song, Zhou, Chen, and Shi}]{lin2023moprd}
Jialiang Lin, Jiaxin Song, Zhangping Zhou, Yidong Chen, and Xiaodong Shi. 2023.
\newblock Moprd: A multidisciplinary open peer review dataset.
\newblock \emph{Neural Computing and Applications}, 35(34):24191--24206.

\bibitem[{OpenAI et~al.(2024)OpenAI, Achiam, Adler, Agarwal, Ahmad, Akkaya, Aleman, Almeida, Altenschmidt, Altman, Anadkat, Avila, Babuschkin, Balaji, Balcom, Baltescu, Bao, Bavarian, Belgum, Bello, Berdine, Bernadett-Shapiro, Berner, Bogdonoff, Boiko, Boyd, Brakman, Brockman, Brooks, Brundage, Button, Cai, Campbell, Cann, Carey, Carlson, Carmichael, Chan, Chang, Chantzis, Chen, Chen, Chen, Chen, Chen, Chess, Cho, Chu, Chung, Cummings, Currier, Dai, Decareaux, Degry, Deutsch, Deville, Dhar, Dohan, Dowling, Dunning, Ecoffet, Eleti, Eloundou, Farhi, Fedus, Felix, Fishman, Forte, Fulford, Gao, Georges, Gibson, Goel, Gogineni, Goh, Gontijo-Lopes, Gordon, Grafstein, Gray, Greene, Gross, Gu, Guo, Hallacy, Han, Harris, He, Heaton, Heidecke, Hesse, Hickey, Hickey, Hoeschele, Houghton, Hsu, Hu, Hu, Huizinga, Jain, Jain, Jang, Jiang, Jiang, Jin, Jin, Jomoto, Jonn, Jun, Kaftan, Łukasz Kaiser, Kamali, Kanitscheider, Keskar, Khan, Kilpatrick, Kim, Kim, Kim, Kirchner, Kiros, Knight, Kokotajlo, Łukasz Kondraciuk,
  Kondrich, Konstantinidis, Kosic, Krueger, Kuo, Lampe, Lan, Lee, Leike, Leung, Levy, Li, Lim, Lin, Lin, Litwin, Lopez, Lowe, Lue, Makanju, Malfacini, Manning, Markov, Markovski, Martin, Mayer, Mayne, McGrew, McKinney, McLeavey, McMillan, McNeil, Medina, Mehta, Menick, Metz, Mishchenko, Mishkin, Monaco, Morikawa, Mossing, Mu, Murati, Murk, Mély, Nair, Nakano, Nayak, Neelakantan, Ngo, Noh, Ouyang, O'Keefe, Pachocki, Paino, Palermo, Pantuliano, Parascandolo, Parish, Parparita, Passos, Pavlov, Peng, Perelman, de~Avila Belbute~Peres, Petrov, de~Oliveira~Pinto, Michael, Pokorny, Pokrass, Pong, Powell, Power, Power, Proehl, Puri, Radford, Rae, Ramesh, Raymond, Real, Rimbach, Ross, Rotsted, Roussez, Ryder, Saltarelli, Sanders, Santurkar, Sastry, Schmidt, Schnurr, Schulman, Selsam, Sheppard, Sherbakov, Shieh, Shoker, Shyam, Sidor, Sigler, Simens, Sitkin, Slama, Sohl, Sokolowsky, Song, Staudacher, Such, Summers, Sutskever, Tang, Tezak, Thompson, Tillet, Tootoonchian, Tseng, Tuggle, Turley, Tworek, Uribe, Vallone,
  Vijayvergiya, Voss, Wainwright, Wang, Wang, Wang, Ward, Wei, Weinmann, Welihinda, Welinder, Weng, Weng, Wiethoff, Willner, Winter, Wolrich, Wong, Workman, Wu, Wu, Wu, Xiao, Xu, Yoo, Yu, Yuan, Zaremba, Zellers, Zhang, Zhang, Zhao, Zheng, Zhuang, Zhuk, and Zoph}]{openai2024gpt4}
OpenAI, Josh Achiam, Steven Adler, Sandhini Agarwal, Lama Ahmad, Ilge Akkaya, Florencia~Leoni Aleman, Diogo Almeida, Janko Altenschmidt, Sam Altman, Shyamal Anadkat, Red Avila, Igor Babuschkin, Suchir Balaji, Valerie Balcom, Paul Baltescu, Haiming Bao, Mohammad Bavarian, Jeff Belgum, Irwan Bello, Jake Berdine, Gabriel Bernadett-Shapiro, Christopher Berner, Lenny Bogdonoff, Oleg Boiko, Madelaine Boyd, Anna-Luisa Brakman, Greg Brockman, Tim Brooks, Miles Brundage, Kevin Button, Trevor Cai, Rosie Campbell, Andrew Cann, Brittany Carey, Chelsea Carlson, Rory Carmichael, Brooke Chan, Che Chang, Fotis Chantzis, Derek Chen, Sully Chen, Ruby Chen, Jason Chen, Mark Chen, Ben Chess, Chester Cho, Casey Chu, Hyung~Won Chung, Dave Cummings, Jeremiah Currier, Yunxing Dai, Cory Decareaux, Thomas Degry, Noah Deutsch, Damien Deville, Arka Dhar, David Dohan, Steve Dowling, Sheila Dunning, Adrien Ecoffet, Atty Eleti, Tyna Eloundou, David Farhi, Liam Fedus, Niko Felix, Simón~Posada Fishman, Juston Forte, Isabella Fulford, Leo
  Gao, Elie Georges, Christian Gibson, Vik Goel, Tarun Gogineni, Gabriel Goh, Rapha Gontijo-Lopes, Jonathan Gordon, Morgan Grafstein, Scott Gray, Ryan Greene, Joshua Gross, Shixiang~Shane Gu, Yufei Guo, Chris Hallacy, Jesse Han, Jeff Harris, Yuchen He, Mike Heaton, Johannes Heidecke, Chris Hesse, Alan Hickey, Wade Hickey, Peter Hoeschele, Brandon Houghton, Kenny Hsu, Shengli Hu, Xin Hu, Joost Huizinga, Shantanu Jain, Shawn Jain, Joanne Jang, Angela Jiang, Roger Jiang, Haozhun Jin, Denny Jin, Shino Jomoto, Billie Jonn, Heewoo Jun, Tomer Kaftan, Łukasz Kaiser, Ali Kamali, Ingmar Kanitscheider, Nitish~Shirish Keskar, Tabarak Khan, Logan Kilpatrick, Jong~Wook Kim, Christina Kim, Yongjik Kim, Jan~Hendrik Kirchner, Jamie Kiros, Matt Knight, Daniel Kokotajlo, Łukasz Kondraciuk, Andrew Kondrich, Aris Konstantinidis, Kyle Kosic, Gretchen Krueger, Vishal Kuo, Michael Lampe, Ikai Lan, Teddy Lee, Jan Leike, Jade Leung, Daniel Levy, Chak~Ming Li, Rachel Lim, Molly Lin, Stephanie Lin, Mateusz Litwin, Theresa Lopez, Ryan
  Lowe, Patricia Lue, Anna Makanju, Kim Malfacini, Sam Manning, Todor Markov, Yaniv Markovski, Bianca Martin, Katie Mayer, Andrew Mayne, Bob McGrew, Scott~Mayer McKinney, Christine McLeavey, Paul McMillan, Jake McNeil, David Medina, Aalok Mehta, Jacob Menick, Luke Metz, Andrey Mishchenko, Pamela Mishkin, Vinnie Monaco, Evan Morikawa, Daniel Mossing, Tong Mu, Mira Murati, Oleg Murk, David Mély, Ashvin Nair, Reiichiro Nakano, Rajeev Nayak, Arvind Neelakantan, Richard Ngo, Hyeonwoo Noh, Long Ouyang, Cullen O'Keefe, Jakub Pachocki, Alex Paino, Joe Palermo, Ashley Pantuliano, Giambattista Parascandolo, Joel Parish, Emy Parparita, Alex Passos, Mikhail Pavlov, Andrew Peng, Adam Perelman, Filipe de~Avila Belbute~Peres, Michael Petrov, Henrique~Ponde de~Oliveira~Pinto, Michael, Pokorny, Michelle Pokrass, Vitchyr~H. Pong, Tolly Powell, Alethea Power, Boris Power, Elizabeth Proehl, Raul Puri, Alec Radford, Jack Rae, Aditya Ramesh, Cameron Raymond, Francis Real, Kendra Rimbach, Carl Ross, Bob Rotsted, Henri Roussez,
  Nick Ryder, Mario Saltarelli, Ted Sanders, Shibani Santurkar, Girish Sastry, Heather Schmidt, David Schnurr, John Schulman, Daniel Selsam, Kyla Sheppard, Toki Sherbakov, Jessica Shieh, Sarah Shoker, Pranav Shyam, Szymon Sidor, Eric Sigler, Maddie Simens, Jordan Sitkin, Katarina Slama, Ian Sohl, Benjamin Sokolowsky, Yang Song, Natalie Staudacher, Felipe~Petroski Such, Natalie Summers, Ilya Sutskever, Jie Tang, Nikolas Tezak, Madeleine~B. Thompson, Phil Tillet, Amin Tootoonchian, Elizabeth Tseng, Preston Tuggle, Nick Turley, Jerry Tworek, Juan Felipe~Cerón Uribe, Andrea Vallone, Arun Vijayvergiya, Chelsea Voss, Carroll Wainwright, Justin~Jay Wang, Alvin Wang, Ben Wang, Jonathan Ward, Jason Wei, CJ~Weinmann, Akila Welihinda, Peter Welinder, Jiayi Weng, Lilian Weng, Matt Wiethoff, Dave Willner, Clemens Winter, Samuel Wolrich, Hannah Wong, Lauren Workman, Sherwin Wu, Jeff Wu, Michael Wu, Kai Xiao, Tao Xu, Sarah Yoo, Kevin Yu, Qiming Yuan, Wojciech Zaremba, Rowan Zellers, Chong Zhang, Marvin Zhang, Shengjia
  Zhao, Tianhao Zheng, Juntang Zhuang, William Zhuk, and Barret Zoph. 2024.
\newblock \href {https://arxiv.org/abs/2303.08774} {Gpt-4 technical report}.
\newblock \emph{Preprint}, arXiv:2303.08774.

\bibitem[{Plank and van Dalen(2019)}]{plank2019citetracked}
Barbara Plank and Reinard van Dalen. 2019.
\newblock Citetracked: A longitudinal dataset of peer reviews and citations.
\newblock In \emph{Proceedings of the 4th Joint Workshop on Bibliometric-enhanced Information Retrieval and Natural Language Processing for Digital Libraries (BIRNDL 2019) co-located with the 42nd International ACM SIGIR Conference on Research and Development in Information Retrieval (SIGIR 2019)}, pages 116--122. CEUR Workshop Proceedings (CEUR-WS. org).

\bibitem[{Press et~al.(2022)Press, Smith, and Lewis}]{press2022train}
Ofir Press, Noah~A. Smith, and Mike Lewis. 2022.
\newblock \href {https://arxiv.org/abs/2108.12409} {Train short, test long: Attention with linear biases enables input length extrapolation}.
\newblock \emph{Preprint}, arXiv:2108.12409.

\bibitem[{Rafailov et~al.(2024)Rafailov, Sharma, Mitchell, Manning, Ermon, and Finn}]{rafailov2024direct}
Rafael Rafailov, Archit Sharma, Eric Mitchell, Christopher~D Manning, Stefano Ermon, and Chelsea Finn. 2024.
\newblock Direct preference optimization: Your language model is secretly a reward model.
\newblock \emph{Advances in Neural Information Processing Systems}, 36.

\bibitem[{Stappen et~al.(2020)Stappen, Rizos, Hasan, Hain, and Schuller}]{stappen20_interspeech}
Lukas Stappen, Georgios Rizos, Madina Hasan, Thomas Hain, and Björn~W. Schuller. 2020.
\newblock \href {https://doi.org/10.21437/Interspeech.2020-2862} {{Uncertainty-Aware Machine Support for Paper Reviewing on the Interspeech 2019 Submission Corpus}}.
\newblock In \emph{Proc. Interspeech 2020}, pages 1808--1812.

\bibitem[{Team et~al.(2024)Team, Reid, Savinov, Teplyashin, Dmitry, Lepikhin, Lillicrap, baptiste Alayrac, Soricut, Lazaridou, Firat, Schrittwieser, Antonoglou, Anil, Borgeaud, Dai, Millican, Dyer, Glaese, Sottiaux, Lee, Viola, Reynolds, Xu, Molloy, Chen, Isard, Barham, Hennigan, McIlroy, Johnson, Schalkwyk, Collins, Rutherford, Moreira, Ayoub, Goel, Meyer, Thornton, Yang, Michalewski, Abbas, Schucher, Anand, Ives, Keeling, Lenc, Haykal, Shakeri, Shyam, Chowdhery, Ring, Spencer, Sezener, Vilnis, Chang, Morioka, Tucker, Zheng, Woodman, Attaluri, Kocisky, Eltyshev, Chen, Chung, Selo, Brahma, Georgiev, Slone, Zhu, Lottes, Qiao, Caine, Riedel, Tomala, Chadwick, Love, Choy, Mittal, Houlsby, Tang, Lamm, Bai, Zhang, He, Cheng, Humphreys, Li, Brin, Cassirer, Miao, Zilka, Tobin, Xu, Proleev, Sohn, Magni, Hendricks, Gao, Ontanon, Bunyan, Byrd, Sharma, Zhang, Pinto, Sinha, Mehta, Jia, Caelles, Webson, Morris, Roelofs, Ding, Strudel, Xiong, Ritter, Dehghani, Chaabouni, Karmarkar, Lai, Mentzer, Xu, Li, Zhang, Paine,
  Goldin, Neyshabur, Baumli, Levskaya, Laskin, Jia, Rae, Xiao, He, Giordano, Yagati, Lespiau, Natsev, Ganapathy, Liu, Martins, Chen, Xu, Barnes, May, Vezer, Oh, Franko, Bridgers, Zhao, Wu, Mustafa, Sechrist, Parisotto, Pillai, Larkin, Gu, Sorokin, Krikun, Guseynov, Landon, Datta, Pritzel, Thacker, Yang, Hui, Hauth, Yeh, Barker, Mao-Jones, Austin, Sheahan, Schuh, Svensson, Jain, Ramasesh, Briukhov, Chung, von Glehn, Butterfield, Jhakra, Wiethoff, Frye, Grimstad, Changpinyo, Lan, Bortsova, Wu, Voigtlaender, Sainath, Gu, Smith, Hawkins, Cao, Besley, Srinivasan, Omernick, Gaffney, Surita, Burnell, Damoc, Ahn, Brock, Pajarskas, Petrushkina, Noury, Blanco, Swersky, Ahuja, Avrahami, Misra, de~Liedekerke, Iinuma, Polozov, York, van~den Driessche, Michel, Chiu, Blevins, Gleicher, Recasens, Rrustemi, Gribovskaya, Roy, Gworek, Arnold, Lee, Lee-Thorp, Maggioni, Piqueras, Badola, Vikram, Gonzalez, Baddepudi, Senter, Devlin, Qin, Azzam, Trebacz, Polacek, Krishnakumar, yiin Chang, Tung, Penchev, Joshi, Olszewska, Muir,
  Wirth, Hartman, Newlan, Kashem, Bolina, Dabir, van Amersfoort, Ahmed, Cobon-Kerr, Kamath, Hrafnkelsson, Hou, Mackinnon, Frechette, Noland, Si, Taropa, Li, Crone, Gulati, Cevey, Adler, Ma, Silver, Tokumine, Powell, Lee, Vodrahalli, Hassan, Mincu, Yang, Levine, Brennan, Wang, Hodkinson, Zhao, Lipschultz, Pope, Chang, Li, Shafey, Paganini, Douglas, Bohnet, Pardo, Odoom, Rosca, dos Santos, Soparkar, Guez, Hudson, Hansen, Asawaroengchai, Addanki, Yu, Stokowiec, Khan, Gilmer, Lee, Bostock, Rong, Caton, Pejman, Pavetic, Brown, Sharma, Lučić, Samuel, Djolonga, Mandhane, Sjösund, Buchatskaya, White, Clay, Jiang, Lim, Hemsley, Cankara, Labanowski, Cao, Steiner, Hashemi, Austin, Gergely, Blyth, Stanton, Shivakumar, Siddhant, Andreassen, Araya, Sethi, Shivanna, Hand, Bapna, Khodaei, Miech, Tanzer, Swing, Thakoor, Aroyo, Pan, Nado, Sygnowski, Winkler, Yu, Saleh, Maggiore, Bansal, Garcia, Kazemi, Patil, Dasgupta, Barr, Giang, Kagohara, Danihelka, Marathe, Feinberg, Elhawaty, Ghelani, Horgan, Miller, Walker, Tanburn,
  Tariq, Shrivastava, Xia, Wang, Chiu, Ashwood, Baatarsukh, Samangooei, Kaufman, Alcober, Stjerngren, Komarek, Tsihlas, Boral, Comanescu, Chen, Liu, Welty, Bloxwich, Chen, Sun, Feng, Mauger, Dotiwalla, Hellendoorn, Sharman, Zheng, Haridasan, Barth-Maron, Swanson, Rogozińska, Andreev, Rubenstein, Sang, Hurt, Elsayed, Wang, Lacey, Ilić, Zhao, Iwanicki, Lince, Chen, Lyu, Lebsack, Griffith, Gaba, Sandhu, Chen, Koop, Rajwar, Yeganeh, Chang, Zhu, Radpour, Davoodi, Lei, Xu, Toyama, Segal, Wicke, Lin, Bulanova, Badia, Rakićević, Sprechmann, Filos, Hou, Campos, Kassner, Sachan, Fortunato, Iwuanyanwu, Nikolaev, Lakshminarayanan, Jazayeri, Varadarajan, Tekur, Fritz, Khalman, Reitter, Dasgupta, Sarcar, Ornduff, Snaider, Huot, Jia, Kemp, Trdin, Vijayakumar, Kim, Angermueller, Lao, Liu, Zhang, Engel, Greene, White, Austin, Taylor, Ashraf, Liu, Georgaki, Cai, Kulizhskaya, Goenka, Saeta, Xu, Frank, de~Cesare, Robenek, Richardson, Alnahlawi, Yew, Ponnapalli, Tagliasacchi, Korchemniy, Kim, Li, Rosgen, Levin, Wiesner,
  Banzal, Srinivasan, Yu, Çağlar Ünlü, Reid, Tung, Finchelstein, Kumar, Elisseeff, Huang, Zhang, Aguilar, Giménez, Xia, Dousse, Gierke, Yates, Jalan, Li, Latorre-Chimoto, Nguyen, Durden, Kallakuri, Liu, Johnson, Tsai, Talbert, Liu, Neitz, Elkind, Selvi, Jasarevic, Soares, Cui, Wang, Wang, Ye, Kallarackal, Loher, Lam, Broder, Holtmann-Rice, Martin, Ramadhana, Shukla, Basu, Mohan, Fernando, Fiedel, Paterson, Li, Garg, Park, Choi, Wu, Singh, Zhang, Globerson, Yu, Carpenter, de~Chaumont~Quitry, Radebaugh, Lin, Tudor, Shroff, Garmon, Du, Vats, Lu, Iqbal, Yakubovich, Tripuraneni, Manyika, Qureshi, Hua, Ngani, Raad, Forbes, Stanway, Sundararajan, Ungureanu, Bishop, Li, Venkatraman, Li, Thornton, Scellato, Gupta, Wang, Tenney, Wu, Shenoy, Carvajal, Wright, Bariach, Xiao, Hawkins, Dalmia, Farabet, Valenzuela, Yuan, Agarwal, Chen, Kim, Hulse, Dukkipati, Paszke, Bolt, Choo, Beattie, Prendki, Vashisht, Santamaria-Fernandez, Cobo, Wilkiewicz, Madras, Elqursh, Uy, Ramirez, Harvey, Liechty, Zen, Seibert, Hu, Khorlin,
  Le, Aharoni, Li, Wang, Kumar, Casagrande, Hoover, Badawy, Soergel, Vnukov, Miecnikowski, Simsa, Kumar, Sellam, Vlasic, Daruki, Shabat, Zhang, Su, Zhang, Liu, Sun, Palmer, Ghaffarkhah, Xiong, Cotruta, Fink, Dixon, Sreevatsa, Goedeckemeyer, Dimitriev, Jafari, Crocker, FitzGerald, Kumar, Ghemawat, Philips, Liu, Liang, Sterneck, Repina, Wu, Knight, Georgiev, Lee, Askham, Chakladar, Louis, Crous, Cate, Petrova, Quinn, Owusu-Afriyie, Singhal, Wei, Kim, Vincent, Nasr, Choquette-Choo, Tojo, Lu, de~Las~Casas, Cheng, Bolukbasi, Lee, Fatehi, Ananthanarayanan, Patel, Kaed, Li, Belle, Chen, Konzelmann, Põder, Garg, Koverkathu, Brown, Dyer, Liu, Nova, Xu, Walton, Parrish, Epstein, McCarthy, Petrov, Hassabis, Kavukcuoglu, Dean, and Vinyals}]{geminiteam2024gemini}
Gemini Team, Machel Reid, Nikolay Savinov, Denis Teplyashin, Dmitry, Lepikhin, Timothy Lillicrap, Jean baptiste Alayrac, Radu Soricut, Angeliki Lazaridou, Orhan Firat, Julian Schrittwieser, Ioannis Antonoglou, Rohan Anil, Sebastian Borgeaud, Andrew Dai, Katie Millican, Ethan Dyer, Mia Glaese, Thibault Sottiaux, Benjamin Lee, Fabio Viola, Malcolm Reynolds, Yuanzhong Xu, James Molloy, Jilin Chen, Michael Isard, Paul Barham, Tom Hennigan, Ross McIlroy, Melvin Johnson, Johan Schalkwyk, Eli Collins, Eliza Rutherford, Erica Moreira, Kareem Ayoub, Megha Goel, Clemens Meyer, Gregory Thornton, Zhen Yang, Henryk Michalewski, Zaheer Abbas, Nathan Schucher, Ankesh Anand, Richard Ives, James Keeling, Karel Lenc, Salem Haykal, Siamak Shakeri, Pranav Shyam, Aakanksha Chowdhery, Roman Ring, Stephen Spencer, Eren Sezener, Luke Vilnis, Oscar Chang, Nobuyuki Morioka, George Tucker, Ce~Zheng, Oliver Woodman, Nithya Attaluri, Tomas Kocisky, Evgenii Eltyshev, Xi~Chen, Timothy Chung, Vittorio Selo, Siddhartha Brahma, Petko
  Georgiev, Ambrose Slone, Zhenkai Zhu, James Lottes, Siyuan Qiao, Ben Caine, Sebastian Riedel, Alex Tomala, Martin Chadwick, Juliette Love, Peter Choy, Sid Mittal, Neil Houlsby, Yunhao Tang, Matthew Lamm, Libin Bai, Qiao Zhang, Luheng He, Yong Cheng, Peter Humphreys, Yujia Li, Sergey Brin, Albin Cassirer, Yingjie Miao, Lukas Zilka, Taylor Tobin, Kelvin Xu, Lev Proleev, Daniel Sohn, Alberto Magni, Lisa~Anne Hendricks, Isabel Gao, Santiago Ontanon, Oskar Bunyan, Nathan Byrd, Abhanshu Sharma, Biao Zhang, Mario Pinto, Rishika Sinha, Harsh Mehta, Dawei Jia, Sergi Caelles, Albert Webson, Alex Morris, Becca Roelofs, Yifan Ding, Robin Strudel, Xuehan Xiong, Marvin Ritter, Mostafa Dehghani, Rahma Chaabouni, Abhijit Karmarkar, Guangda Lai, Fabian Mentzer, Bibo Xu, YaGuang Li, Yujing Zhang, Tom~Le Paine, Alex Goldin, Behnam Neyshabur, Kate Baumli, Anselm Levskaya, Michael Laskin, Wenhao Jia, Jack~W. Rae, Kefan Xiao, Antoine He, Skye Giordano, Lakshman Yagati, Jean-Baptiste Lespiau, Paul Natsev, Sanjay Ganapathy, Fangyu
  Liu, Danilo Martins, Nanxin Chen, Yunhan Xu, Megan Barnes, Rhys May, Arpi Vezer, Junhyuk Oh, Ken Franko, Sophie Bridgers, Ruizhe Zhao, Boxi Wu, Basil Mustafa, Sean Sechrist, Emilio Parisotto, Thanumalayan~Sankaranarayana Pillai, Chris Larkin, Chenjie Gu, Christina Sorokin, Maxim Krikun, Alexey Guseynov, Jessica Landon, Romina Datta, Alexander Pritzel, Phoebe Thacker, Fan Yang, Kevin Hui, Anja Hauth, Chih-Kuan Yeh, David Barker, Justin Mao-Jones, Sophia Austin, Hannah Sheahan, Parker Schuh, James Svensson, Rohan Jain, Vinay Ramasesh, Anton Briukhov, Da-Woon Chung, Tamara von Glehn, Christina Butterfield, Priya Jhakra, Matthew Wiethoff, Justin Frye, Jordan Grimstad, Beer Changpinyo, Charline~Le Lan, Anna Bortsova, Yonghui Wu, Paul Voigtlaender, Tara Sainath, Shane Gu, Charlotte Smith, Will Hawkins, Kris Cao, James Besley, Srivatsan Srinivasan, Mark Omernick, Colin Gaffney, Gabriela Surita, Ryan Burnell, Bogdan Damoc, Junwhan Ahn, Andrew Brock, Mantas Pajarskas, Anastasia Petrushkina, Seb Noury, Lorenzo
  Blanco, Kevin Swersky, Arun Ahuja, Thi Avrahami, Vedant Misra, Raoul de~Liedekerke, Mariko Iinuma, Alex Polozov, Sarah York, George van~den Driessche, Paul Michel, Justin Chiu, Rory Blevins, Zach Gleicher, Adrià Recasens, Alban Rrustemi, Elena Gribovskaya, Aurko Roy, Wiktor Gworek, Sébastien M.~R. Arnold, Lisa Lee, James Lee-Thorp, Marcello Maggioni, Enrique Piqueras, Kartikeya Badola, Sharad Vikram, Lucas Gonzalez, Anirudh Baddepudi, Evan Senter, Jacob Devlin, James Qin, Michael Azzam, Maja Trebacz, Martin Polacek, Kashyap Krishnakumar, Shuo yiin Chang, Matthew Tung, Ivo Penchev, Rishabh Joshi, Kate Olszewska, Carrie Muir, Mateo Wirth, Ale~Jakse Hartman, Josh Newlan, Sheleem Kashem, Vijay Bolina, Elahe Dabir, Joost van Amersfoort, Zafarali Ahmed, James Cobon-Kerr, Aishwarya Kamath, Arnar~Mar Hrafnkelsson, Le~Hou, Ian Mackinnon, Alexandre Frechette, Eric Noland, Xiance Si, Emanuel Taropa, Dong Li, Phil Crone, Anmol Gulati, Sébastien Cevey, Jonas Adler, Ada Ma, David Silver, Simon Tokumine, Richard
  Powell, Stephan Lee, Kiran Vodrahalli, Samer Hassan, Diana Mincu, Antoine Yang, Nir Levine, Jenny Brennan, Mingqiu Wang, Sarah Hodkinson, Jeffrey Zhao, Josh Lipschultz, Aedan Pope, Michael~B. Chang, Cheng Li, Laurent~El Shafey, Michela Paganini, Sholto Douglas, Bernd Bohnet, Fabio Pardo, Seth Odoom, Mihaela Rosca, Cicero~Nogueira dos Santos, Kedar Soparkar, Arthur Guez, Tom Hudson, Steven Hansen, Chulayuth Asawaroengchai, Ravi Addanki, Tianhe Yu, Wojciech Stokowiec, Mina Khan, Justin Gilmer, Jaehoon Lee, Carrie~Grimes Bostock, Keran Rong, Jonathan Caton, Pedram Pejman, Filip Pavetic, Geoff Brown, Vivek Sharma, Mario Lučić, Rajkumar Samuel, Josip Djolonga, Amol Mandhane, Lars~Lowe Sjösund, Elena Buchatskaya, Elspeth White, Natalie Clay, Jiepu Jiang, Hyeontaek Lim, Ross Hemsley, Zeyncep Cankara, Jane Labanowski, Nicola~De Cao, David Steiner, Sayed~Hadi Hashemi, Jacob Austin, Anita Gergely, Tim Blyth, Joe Stanton, Kaushik Shivakumar, Aditya Siddhant, Anders Andreassen, Carlos Araya, Nikhil Sethi, Rakesh
  Shivanna, Steven Hand, Ankur Bapna, Ali Khodaei, Antoine Miech, Garrett Tanzer, Andy Swing, Shantanu Thakoor, Lora Aroyo, Zhufeng Pan, Zachary Nado, Jakub Sygnowski, Stephanie Winkler, Dian Yu, Mohammad Saleh, Loren Maggiore, Yamini Bansal, Xavier Garcia, Mehran Kazemi, Piyush Patil, Ishita Dasgupta, Iain Barr, Minh Giang, Thais Kagohara, Ivo Danihelka, Amit Marathe, Vladimir Feinberg, Mohamed Elhawaty, Nimesh Ghelani, Dan Horgan, Helen Miller, Lexi Walker, Richard Tanburn, Mukarram Tariq, Disha Shrivastava, Fei Xia, Qingze Wang, Chung-Cheng Chiu, Zoe Ashwood, Khuslen Baatarsukh, Sina Samangooei, Raphaël~Lopez Kaufman, Fred Alcober, Axel Stjerngren, Paul Komarek, Katerina Tsihlas, Anudhyan Boral, Ramona Comanescu, Jeremy Chen, Ruibo Liu, Chris Welty, Dawn Bloxwich, Charlie Chen, Yanhua Sun, Fangxiaoyu Feng, Matthew Mauger, Xerxes Dotiwalla, Vincent Hellendoorn, Michael Sharman, Ivy Zheng, Krishna Haridasan, Gabe Barth-Maron, Craig Swanson, Dominika Rogozińska, Alek Andreev, Paul~Kishan Rubenstein, Ruoxin
  Sang, Dan Hurt, Gamaleldin Elsayed, Renshen Wang, Dave Lacey, Anastasija Ilić, Yao Zhao, Adam Iwanicki, Alejandro Lince, Alexander Chen, Christina Lyu, Carl Lebsack, Jordan Griffith, Meenu Gaba, Paramjit Sandhu, Phil Chen, Anna Koop, Ravi Rajwar, Soheil~Hassas Yeganeh, Solomon Chang, Rui Zhu, Soroush Radpour, Elnaz Davoodi, Ving~Ian Lei, Yang Xu, Daniel Toyama, Constant Segal, Martin Wicke, Hanzhao Lin, Anna Bulanova, Adrià~Puigdomènech Badia, Nemanja Rakićević, Pablo Sprechmann, Angelos Filos, Shaobo Hou, Víctor Campos, Nora Kassner, Devendra Sachan, Meire Fortunato, Chimezie Iwuanyanwu, Vitaly Nikolaev, Balaji Lakshminarayanan, Sadegh Jazayeri, Mani Varadarajan, Chetan Tekur, Doug Fritz, Misha Khalman, David Reitter, Kingshuk Dasgupta, Shourya Sarcar, Tina Ornduff, Javier Snaider, Fantine Huot, Johnson Jia, Rupert Kemp, Nejc Trdin, Anitha Vijayakumar, Lucy Kim, Christof Angermueller, Li~Lao, Tianqi Liu, Haibin Zhang, David Engel, Somer Greene, Anaïs White, Jessica Austin, Lilly Taylor, Shereen
  Ashraf, Dangyi Liu, Maria Georgaki, Irene Cai, Yana Kulizhskaya, Sonam Goenka, Brennan Saeta, Ying Xu, Christian Frank, Dario de~Cesare, Brona Robenek, Harry Richardson, Mahmoud Alnahlawi, Christopher Yew, Priya Ponnapalli, Marco Tagliasacchi, Alex Korchemniy, Yelin Kim, Dinghua Li, Bill Rosgen, Kyle Levin, Jeremy Wiesner, Praseem Banzal, Praveen Srinivasan, Hongkun Yu, Çağlar Ünlü, David Reid, Zora Tung, Daniel Finchelstein, Ravin Kumar, Andre Elisseeff, Jin Huang, Ming Zhang, Ricardo Aguilar, Mai Giménez, Jiawei Xia, Olivier Dousse, Willi Gierke, Damion Yates, Komal Jalan, Lu~Li, Eri Latorre-Chimoto, Duc~Dung Nguyen, Ken Durden, Praveen Kallakuri, Yaxin Liu, Matthew Johnson, Tomy Tsai, Alice Talbert, Jasmine Liu, Alexander Neitz, Chen Elkind, Marco Selvi, Mimi Jasarevic, Livio~Baldini Soares, Albert Cui, Pidong Wang, Alek~Wenjiao Wang, Xinyu Ye, Krystal Kallarackal, Lucia Loher, Hoi Lam, Josef Broder, Dan Holtmann-Rice, Nina Martin, Bramandia Ramadhana, Mrinal Shukla, Sujoy Basu, Abhi Mohan, Nick
  Fernando, Noah Fiedel, Kim Paterson, Hui Li, Ankush Garg, Jane Park, DongHyun Choi, Diane Wu, Sankalp Singh, Zhishuai Zhang, Amir Globerson, Lily Yu, John Carpenter, Félix de~Chaumont~Quitry, Carey Radebaugh, Chu-Cheng Lin, Alex Tudor, Prakash Shroff, Drew Garmon, Dayou Du, Neera Vats, Han Lu, Shariq Iqbal, Alex Yakubovich, Nilesh Tripuraneni, James Manyika, Haroon Qureshi, Nan Hua, Christel Ngani, Maria~Abi Raad, Hannah Forbes, Jeff Stanway, Mukund Sundararajan, Victor Ungureanu, Colton Bishop, Yunjie Li, Balaji Venkatraman, Bo~Li, Chloe Thornton, Salvatore Scellato, Nishesh Gupta, Yicheng Wang, Ian Tenney, Xihui Wu, Ashish Shenoy, Gabriel Carvajal, Diana~Gage Wright, Ben Bariach, Zhuyun Xiao, Peter Hawkins, Sid Dalmia, Clement Farabet, Pedro Valenzuela, Quan Yuan, Ananth Agarwal, Mia Chen, Wooyeol Kim, Brice Hulse, Nandita Dukkipati, Adam Paszke, Andrew Bolt, Kiam Choo, Jennifer Beattie, Jennifer Prendki, Harsha Vashisht, Rebeca Santamaria-Fernandez, Luis~C. Cobo, Jarek Wilkiewicz, David Madras, Ali
  Elqursh, Grant Uy, Kevin Ramirez, Matt Harvey, Tyler Liechty, Heiga Zen, Jeff Seibert, Clara~Huiyi Hu, Andrey Khorlin, Maigo Le, Asaf Aharoni, Megan Li, Lily Wang, Sandeep Kumar, Norman Casagrande, Jay Hoover, Dalia~El Badawy, David Soergel, Denis Vnukov, Matt Miecnikowski, Jiri Simsa, Praveen Kumar, Thibault Sellam, Daniel Vlasic, Samira Daruki, Nir Shabat, John Zhang, Guolong Su, Jiageng Zhang, Jeremiah Liu, Yi~Sun, Evan Palmer, Alireza Ghaffarkhah, Xi~Xiong, Victor Cotruta, Michael Fink, Lucas Dixon, Ashwin Sreevatsa, Adrian Goedeckemeyer, Alek Dimitriev, Mohsen Jafari, Remi Crocker, Nicholas FitzGerald, Aviral Kumar, Sanjay Ghemawat, Ivan Philips, Frederick Liu, Yannie Liang, Rachel Sterneck, Alena Repina, Marcus Wu, Laura Knight, Marin Georgiev, Hyo Lee, Harry Askham, Abhishek Chakladar, Annie Louis, Carl Crous, Hardie Cate, Dessie Petrova, Michael Quinn, Denese Owusu-Afriyie, Achintya Singhal, Nan Wei, Solomon Kim, Damien Vincent, Milad Nasr, Christopher~A. Choquette-Choo, Reiko Tojo, Shawn Lu, Diego
  de~Las~Casas, Yuchung Cheng, Tolga Bolukbasi, Katherine Lee, Saaber Fatehi, Rajagopal Ananthanarayanan, Miteyan Patel, Charbel Kaed, Jing Li, Shreyas~Rammohan Belle, Zhe Chen, Jaclyn Konzelmann, Siim Põder, Roopal Garg, Vinod Koverkathu, Adam Brown, Chris Dyer, Rosanne Liu, Azade Nova, Jun Xu, Alanna Walton, Alicia Parrish, Mark Epstein, Sara McCarthy, Slav Petrov, Demis Hassabis, Koray Kavukcuoglu, Jeffrey Dean, and Oriol Vinyals. 2024.
\newblock \href {https://arxiv.org/abs/2403.05530} {Gemini 1.5: Unlocking multimodal understanding across millions of tokens of context}.
\newblock \emph{Preprint}, arXiv:2403.05530.

\bibitem[{Tian et~al.(2024)Tian, Zhao, Wang, and Ding}]{tian2024mmrec}
Jiahao Tian, Jinman Zhao, Zhenkai Wang, and Zhicheng Ding. 2024.
\newblock Mmrec: Llm based multi-modal recommender system.
\newblock \emph{arXiv preprint arXiv:2408.04211}.

\bibitem[{Touvron et~al.(2023)Touvron, Martin, Stone, Albert, Almahairi, Babaei, Bashlykov, Batra, Bhargava, Bhosale et~al.}]{touvron2023llama}
Hugo Touvron, Louis Martin, Kevin Stone, Peter Albert, Amjad Almahairi, Yasmine Babaei, Nikolay Bashlykov, Soumya Batra, Prajjwal Bhargava, Shruti Bhosale, et~al. 2023.
\newblock Llama 2: Open foundation and fine-tuned chat models.
\newblock \emph{arXiv preprint arXiv:2307.09288}.

\bibitem[{Wang et~al.(2023)Wang, Dong, Cheng, Liu, Yan, Gao, and Wei}]{wang2023augmenting}
Weizhi Wang, Li~Dong, Hao Cheng, Xiaodong Liu, Xifeng Yan, Jianfeng Gao, and Furu Wei. 2023.
\newblock \href {https://arxiv.org/abs/2306.07174} {Augmenting language models with long-term memory}.
\newblock \emph{Preprint}, arXiv:2306.07174.

\bibitem[{Wu et~al.(2022)Wu, Lan, Qian, Gu, Geramifard, and Yu}]{wu2022memformer}
Qingyang Wu, Zhenzhong Lan, Kun Qian, Jing Gu, Alborz Geramifard, and Zhou Yu. 2022.
\newblock \href {https://arxiv.org/abs/2010.06891} {Memformer: A memory-augmented transformer for sequence modeling}.
\newblock \emph{Preprint}, arXiv:2010.06891.

\bibitem[{Xiong et~al.(2023)Xiong, Liu, Molybog, Zhang, Bhargava, Hou, Martin, Rungta, Sankararaman, Oguz et~al.}]{xiong2023effective}
Wenhan Xiong, Jingyu Liu, Igor Molybog, Hejia Zhang, Prajjwal Bhargava, Rui Hou, Louis Martin, Rashi Rungta, Karthik~Abinav Sankararaman, Barlas Oguz, et~al. 2023.
\newblock Effective long-context scaling of foundation models.
\newblock \emph{arXiv preprint arXiv:2309.16039}.

\bibitem[{Zheng et~al.(2023)Zheng, Chiang, Sheng, Zhuang, Wu, Zhuang, Lin, Li, Li, Xing, Zhang, Gonzalez, and Stoica}]{zheng2023judging}
Lianmin Zheng, Wei-Lin Chiang, Ying Sheng, Siyuan Zhuang, Zhanghao Wu, Yonghao Zhuang, Zi~Lin, Zhuohan Li, Dacheng Li, Eric~P. Xing, Hao Zhang, Joseph~E. Gonzalez, and Ion Stoica. 2023.
\newblock \href {https://arxiv.org/abs/2306.05685} {Judging llm-as-a-judge with mt-bench and chatbot arena}.
\newblock \emph{Preprint}, arXiv:2306.05685.

\end{thebibliography}

\appendix
\section{Hyperparameters}
The detailed information can be found in Table \ref{tab:Modelparameters}. 
\begin{table*}[ht]
\centering
\caption{Model parameters and hyperparameters setup for reproduction. The base model is microsoft/Phi-3-mini-128k-instruct. The training was conducted using Deepspeed Stage-3 on a 4x A100 80GB GPU machine with LoRA for parameter-efficient fine-tuning. DPO and SFT (including SteerLM and rejection sampling) employed distinct hyperparameters}
\begin{tabular}{lll}
\hline
\textbf{Parameter Settings} & \textbf{Name} & \textbf{Value} \\ \hline
Lora Setting & Lora Rank & 128 \\
 & Lora Alpha & 256 \\
DPO Setting & Model & microsoft/Phi-3-mini-128k-instruct \\
 & Flash Attention & fa2 \\
 & Do Train & true \\
 & Fine-tuning Type & lora \\
 & Pref Beta & 0.1 \\
 & Pref Loss & sigmoid \\
 & Cutoff Length & 32000 \\
 & Overwrite Cache & true \\
 & Logging Steps & 1 \\
 & Save Steps & 500 \\
 & Plot Loss & true \\
 & Overwrite Output Directory & true \\
 & Train Batch Size & 1 \\
 & Gradient Accumulation Steps & 16 \\
 & Learning Rate & 5.0e-6 \\
 & Num Train Epochs & 3.0 \\
 & LR Scheduler Type & cosine \\
 & Warmup Ratio & 0.1 \\
 & FP16 & true \\

SFT Setting & Model & microsoft/Phi-3-mini-128k-instruct \\
 & Fine-tuning Type & lora \\
 & Template & phi \\
 & Cutoff Length & 31000 \\
 & Overwrite Output Directory & true \\
 & Train Batch Size & 2 \\
 & Gradient Accumulation Steps & 2 \\
 & Learning Rate & 1.0e-4 \\
 & Num Train Epochs & 3.0 \\
 & LR Scheduler Type & cosine \\
 & Warmup Ratio & 0.1 \\
 & FP16 & true \\

Common settings & Hardware & 4X A100 80G \\
 & Distributed Learning & Zero-3 \\
 & Use BF-16 & True \\
 & Learning Rate Scheduler & Cosine \\
 & Learning Rate Warm up & 0.003 \\
 & Weight Decay & False \\ \hline
\end{tabular}
\label{tab:Modelparameters}
\end{table*}

\section{Benchmark descriptions}
\textbf{QASPER:} QASPER is a benchmark designed to assess the ability of Question Answering (QA) systems to handle complex reasoning about claims made in multiple parts of academic papers. It comprises 5,049 questions over 1,585 Natural Language Processing papers. Each question is formulated by an NLP practitioner who reads only the title and abstract of the corresponding paper, seeking information present in the full text. The questions are answered by a separate set of NLP practitioners who also provide supporting evidence for their answers. We utilized the QASPER benchmark to evaluate our models, highlighting the challenge of document-grounded, information-seeking QA and motivating further research in this area. 

\textbf{LongBench:} LongBench is the first benchmark for bilingual, multitasking, and comprehensive assessment of long context understanding capabilities of large language models. It includes different languages (Chinese and English) to provide a more comprehensive evaluation of the models' multilingual capabilities in long contexts. Composed of six major categories and twenty-one different tasks, LongBench covers key long-text application scenarios such as single-document QA, multi-document QA, summarization, few-shot learning, synthetic tasks, and code completion. We used LongBench to evaluate our models' performance across these diverse and challenging tasks.
% \section{Additional methods}
% \label{sec:appendix}
% \begin{table}[!h]
% \centering
% \caption{Additional for improving the long-text reading ability of large language models}

% \begin{tabular}{cc}
% \toprule
% \textbf{Research} & \textbf{Author} \\
% \midrule
% Longformer & \cite{beltagy2020longformer} \\ 
% FlashAttention & \cite{dao2022flashattention} \\ 
% Transformer-XL & \cite{dai2019transformerxl} \\ 
% RWKV & \cite{peng2023rwkv} \\ 
% Mamba & \cite{gu2024mamba}\\ 
% LongLoRA & \cite{chen2024longlora}  \\ 
% BGE Landmark Embedding & \cite{luo2024bge}  \\
% LongLLMLingua & \cite{jiang2023longllmlingua}  \\
% \bottomrule
% \end{tabular}
% \label{tab:accents}
% \end{table}
\section{Comparison of Textract, Unstructured.io}\label{sec:appendix-compare-tools}
Unstructured.io provides a more polished output format compared to Textract, which can significantly reduce the time required for post-processing the extraction results. However, as previously mentioned, we faced performance challenges when using Unstructured.io to process our dataset. To illustrate the difference, here is an example of the output generated by Textract:
\begin{lstlisting}[language=json]
{"DocumentMetadata": {"Pages": 1},
 "Blocks": [
  ...
  {"BlockType": "LINE",
   "Confidence": 99.90467834472656,
   "Text": "We propose a novel perspective of viewing large pretrained models as search en-",
   "Geometry": {
    "BoundingBox": {
      "Width": 0.5294037461280823,
      "Height": 0.012670669704675674,
      "Left": 0.23449814319610596,
      "Top": 0.2926112413406372},
    "Polygon": [
      {"X": 0.23449814319610596, "Y": 0.2929380238056183},
      {"X": 0.7638859152793884, "Y": 0.2926112413406372},
      {"X": 0.7639018893241882, "Y": 0.30495572090148926},
      {"X": 0.23451094329357147, "Y": 0.3052819073200226}]
    },
   "Id": "8e6a9e61-a171-4480-ae35-de9a1e7b1f6c",
   "Relationships": [
    {"Type": "CHILD",
     "Ids": [
        "c663961c-381d-496c-a19c-bf72771ea598",
        ...]}]},
  {"BlockType": "LINE",
   "Confidence": 99.8708267211914,
   "Text": "gines, thereby enabling the repurposing of techniques previously used to enhance",
   "Geometry": {...},
   "Id": "22ebd7a4-aa4b-4721-aebf-8764abbb563c",
   "Relationships": [{"Type": "CHILD",
     "Ids": [
      "3670803f-79be-46f3-b3b5-e6cb0e4e384e",
      ...]}]},
  ...
 ]
}
\end{lstlisting}
Textract's output breaks the text into individual lines, which can be challenging to work with. In contrast, Unstructured.io groups the extracted text into a coherent blob and assigns a type label to each element, as displayed below:
\begin{lstlisting}[language=json]
[
  ...
  {
    "type": "NarrativeText",
    "element_id": "0fab02a1005a99b6c1bac2b553e0b58a",
    "text": "We propose a novel perspective ... pretrained models.",
    "metadata": {
      "coordinates": {
        "points": <bounding box>,
        "system": "PixelSpace",
        "layout_width": 1700,
        "layout_height": 2200
      },
      "filetype": "application/pdf",
      "languages": [
        "eng"
      ],
      "page_number": 1,
      "parent_id": "9225595bc787e96a129cf1c0077c63bf",
      "filename": "1256.pdf"
    }
  },
  ...
]
\end{lstlisting}

The estimates of the processing cost for Unstructured.io, Textract are \$260 and  \$5,000, for 3,500 PDF files with 20 pages per document on average, where each page image size is 1700 x 2200 pixels.
\end{document}